\documentclass[10pt,twocolumn,letterpaper]{article}
\usepackage[pagenumbers]{cvpr}
\usepackage{multicol}
\usepackage{array,tabularx,booktabs}
\usepackage[T1]{fontenc}
\usepackage{tgtermes,tgheros}
\definecolor{cvprblue}{rgb}{0.21,0.49,0.74}
\usepackage[pagebackref,breaklinks,colorlinks,allcolors=cvprblue]{hyperref}
\def\confName{CVPR}\def\confYear{2027}

\newcolumntype{Y}{>{\centering\arraybackslash}X}
\newcommand{\tablenote}[1]{\par\smallskip\noindent{\raggedright\fontsize{8}{10}\selectfont #1\par}}
\title{EgoLAM: Learning Latent Actions from Hand and Head Motion}
\author{Ziyi Ding, Yiwei Wang, Zhongxing Wei, Xianxin Lai, Yechi Liu, Xiaodi Huang, Jinghua Hou, Xiaoquan Sun, Haijier Chen, Zhe Liu, Yinan Chen, Junqiao Zhao, Xiao-Ping Zhang, Jiayu Chen}
\makeatletter
\renewcommand{\@maketitle}{\begin{center}
{\Large\bfseries\@title\par}\vspace{4pt}
{\fontsize{11}{14}\selectfont Ziyi Ding\textsuperscript{1,3,*}, Yiwei Wang\textsuperscript{2,*}, Zhongxing Wei\textsuperscript{3}, Xianxin Lai\textsuperscript{4}, Yechi Liu\textsuperscript{5,9}, Xiaodi Huang\textsuperscript{5}, Jinghua Hou\textsuperscript{4}\\
Xiaoquan Sun\textsuperscript{6}, Haijier Chen\textsuperscript{7}, Zhe Liu\textsuperscript{4}, Yinan Chen\textsuperscript{8}, Junqiao Zhao\textsuperscript{2,\textdagger{}}, Xiao-Ping Zhang\textsuperscript{1,\textdagger{}}, Jiayu Chen\textsuperscript{4,3,\textdagger{}}\par}\vspace{2pt}
{\fontsize{9}{11}\selectfont \textsuperscript{1}Tsinghua University\quad \textsuperscript{2}Tongji University\quad \textsuperscript{3}INFIFORCE\quad \textsuperscript{4}The University of Hong Kong\\ \textsuperscript{5}Institute of Automation, Chinese Academy of Sciences\quad \textsuperscript{6}Huazhong University of Science and Technology\\ \textsuperscript{7}Wuhan University\quad \textsuperscript{8}Zhejiang University\quad \textsuperscript{9}University of Chinese Academy of Sciences\par}
{\fontsize{9}{11}\selectfont * Equal contribution (co-first authors).\quad \textdagger{} Corresponding authors.\par}
{\fontsize{9}{11}\selectfont \href{https://dzyy123.github.io/EgoLAM-Pages/}{\raisebox{-2pt}{\includegraphics[height=10pt]{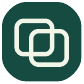}}\hspace{3pt}Project page}\qquad \href{https://github.com/Dzyy123/EgoLAM}{\raisebox{-2pt}{\includegraphics[height=10pt]{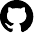}}\hspace{3pt}GitHub code}\qquad \href{https://huggingface.co/dzyy123/EgoLAM}{\raisebox{-2pt}{\includegraphics[height=10pt]{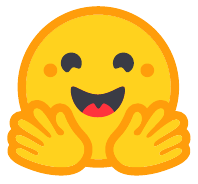}}\hspace{3pt}Hugging Face weights}\par}
\end{center}\vspace{-4pt}}
\makeatother
\hypersetup{pdftitle={EgoLAM: Learning Latent Actions from Hand and Head Motion},pdfsubject={EgoLAM English preprint},pdfauthor={Ziyi Ding, Yiwei Wang, Zhongxing Wei, Xianxin Lai, Yechi Liu, Xiaodi Huang, Jinghua Hou, Xiaoquan Sun, Haijier Chen, Zhe Liu, Yinan Chen, Junqiao Zhao, Xiao-Ping Zhang, Jiayu Chen}}

\begin{document}\maketitle
\setlength{\textfloatsep}{5pt plus 2pt minus 2pt}
\setlength{\dbltextfloatsep}{6pt plus 2pt minus 2pt}
\setlength{\intextsep}{8pt plus 2pt minus 2pt}
\setlength{\floatsep}{8pt plus 2pt minus 2pt}
\captionsetup[figure]{skip=4pt}
\begin{abstract}
Latent action models let a robot policy learn from video by turning the change between two frames into a training target. In egocentric video, a cup can move across the image because a hand pushes it or because the wearer turns. Both changes help predict the next frame, but only the first describes manipulation. During training, EgoLAM supervises the action code with measured wrist motion and supplies camera motion separately to the forward model. Even when the wearer turns while pushing a cup, the training target still describes how the hand pushed it. The action encoder pools object slots around each hand, and wrist translation and rotation can be decoded from the code in a common reference frame. Extracting a code then requires only two RGB frames. Alongside LARYBench classification and regression, we measure how much wrist motion a frozen code retains and how strongly it responds to a synthetic view or lighting change. EgoLAM has the highest hand-motion decoding score among eight evaluated models. Across three LIBERO suites, policies pretrained on its codes and fine-tuned on expert demonstrations achieve 79.38\% mean success, 28.73 points above expert behavior cloning and 9.76 points above the strongest other latent action model in each suite on average.
\end{abstract}

\section{Introduction}
Egocentric videos can teach bimanual manipulation even without recorded robot commands. Policies learn from these videos by predicting latent action labels from observation history. The labels determine which changes a policy learns to imitate.

\begin{figure}[!t]\centering\includegraphics[width=\linewidth]{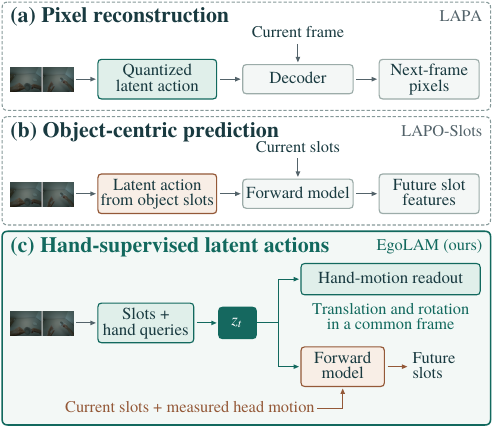}
\caption{\textbf{Three training objectives applied to a cup push and a head turn.} Teal links wrist motion that image reconstruction may omit to EgoLAM's hand supervision. Brown links the apparent displacement of a stationary object to the camera motion supplied to the forward model.}
\label{fig:paradigms}
\end{figure}

\begin{figure*}[t]
\centering\includegraphics[width=\textwidth]{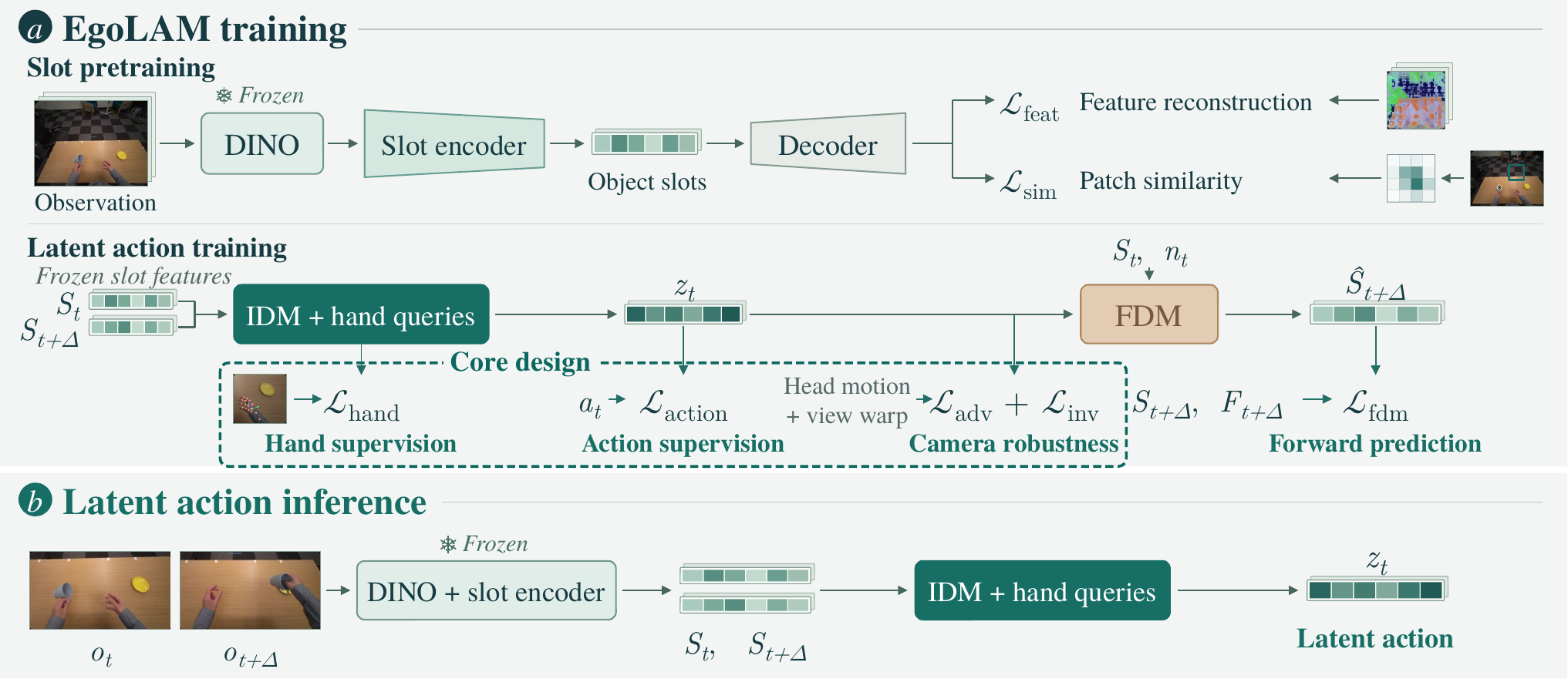}
\caption{\textbf{Three steps from slot pretraining to latent action extraction.} Top: frozen DINO features and cross-frame patch similarities supervise the slots. Middle: the IDM encodes two sets of frozen slots as $z_t$. The dashed box gives the hand and camera constraints. At right, the FDM predicts future slots from the current slots $S_t$, latent action $z_t$, and head motion $n_t$. Bottom: inference uses the RGB pair, frozen front end and IDM. Snowflakes denote frozen parameters.}
\label{fig:overview}
\end{figure*}

\textbf{Motivation.} Applying existing latent action models to egocentric manipulation raises two problems for imitation. \textbf{First, next-frame reconstruction can omit the motion the hand actually performs.} LAPA reconstructs the later frame from the earlier frame and a latent action~\cite{lapa}. During a push, a code may suffice to reconstruct the scene yet omit some wrist translation or rotation. Accurate reconstruction supplies no separate signal to preserve that motion.

\textbf{Second, predicting object slots can still reward encoding a head turn as an action.} LAPO-Slots predicts slots instead of pixels to reduce background distractions~\cite{objectlam}. When the wearer turns left, the cup shifts right in the image even if hand and cup remain still (Figure~\ref{fig:paradigms}). Even the correct slot undergoes this change. Predicting it does not require distinguishing a push from a turn. \textbf{Problem.} How can we train a code to retain the wrist motion of a push without treating a head turn alone as manipulation?

To address both problems, we introduce EgoLAM, a framework that supervises latent actions with wrist motion and supplies head motion separately to the forward model. A query for each hand pools features from eight slots. We express both hands' poses in the earlier head frame, so their translation and rotation describe the push even as the wearer turns. When only the head turns, the hand target is near zero and the forward model can predict the shift from camera motion. Extraction takes just two RGB frames.

This division gives the two annotations distinct roles: the wrists specify what action the code should retain, and the head provides evidence for predicting changes in viewpoint.

\section{Related Work}
\paragraph{Inferring actions from observation transitions.}
A transition between observations can supply an imitation target when expert commands are unavailable. BCO learns inverse dynamics through interaction and infers the demonstrator's actions~\cite{bco}. ILPO learns latent actions first and uses limited interaction to map them to controls~\cite{ilpo}. LAPO learns codes that explain transitions and trains a policy to predict them~\cite{lapo}. For a head-mounted camera, the transition also includes the observer's motion. This makes the choice of what the code must predict particularly important.

\paragraph{Training targets for video action models.}
Pixels, visual features and action labels place different demands on a code. LAPA learns from image reconstruction and Moto tokenizes motion for video pretraining~\cite{lapa,moto}. A change in appearance is informative for visual prediction even when it comes from a head turn. CLAM uses a continuous code with action decoding~\cite{clam}, and LAOM shows how action labels help under visual distraction~\cite{laom}. GeoLAM instead uses a geometry teacher to supervise displacement, residual image motion and orientation changes~\cite{geolam}. EgoLAM specifies whose motion to preserve: its readout target is the wearer's wrist motion expressed in one reference frame. Related approaches use synchronized views~\cite{tcn} or progress toward a goal~\cite{vip} to learn visual representations.

\paragraph{Object-centric dynamics.}
Slots provide a way to follow the manipulated object through a cluttered scene. Slot Attention groups image features~\cite{slotattention}, SAVi tracks slots over time~\cite{savi}, and SlotFormer predicts their dynamics~\cite{slotformer}. DINOSAUR groups self-supervised visual features~\cite{dinosaur,dino}; our temporal training target follows VideoSAUR~\cite{videosaur}. Closest to our work, Klepach et al.~\cite{objectlam} select action-relevant slots and learn from their changes. Selecting the cup still leaves its apparent displacement ambiguous under a moving camera. EgoLAM uses hand queries to pool the interaction and measured camera motion to condition the prediction of future slots.

\paragraph{Evaluating video representations.}
LARYBench measures action classification and robot command regression separately from policy learning~\cite{lary}. Something-Something labels temporal actions~\cite{something}, while Ego4D and Ego-Exo4D cover first-person activities and skills across viewpoints~\cite{ego4d,egoexo}. Recognizing a cup lift says little about its wrist trajectory. Section~\ref{sec:egodiag} therefore tests whether the wrist displacement and rotation can be decoded and how much the code changes when only the image is perturbed.

\section{EgoLAM: Learning Latent Hand Actions from Egocentric Video}
\setlength{\abovedisplayskip}{4pt plus 2pt minus 1pt}
\setlength{\belowdisplayskip}{4pt plus 2pt minus 1pt}
\setlength{\belowdisplayshortskip}{4pt plus 1pt minus 1pt}

\subsection{Problem setup}
Given RGB observations at times $t$ and $t+\Delta$ during an interaction, we seek an action representation
\begin{equation}
z_t=E_\theta(o_t,o_{t+\Delta}),\qquad z_t\in\mathbb R^{d_z}.
\end{equation}
where $\Delta$ is the time between frames and $d_z$ is the latent action dimension. We use 128 in the main model and compare 16, 32, 64, 128, 256 and 512 in Supplementary Table~S3. During training, direct hand supervision requires annotated keypoints or poses (end-effector poses for robots), and camera supervision uses annotated head or camera poses where available. Each supervised loss uses only the examples with its required labels. At inference the only input is the RGB frame pair.

We test three properties: how well hand motion can be decoded, how much a controlled visual change alters the code for the same action, and how often a policy trained on the code completes a task. The first two tests use a frozen model. The third uses closed-loop rollouts.

\subsection{Overview of EgoLAM}\label{sec:method-overview}
\textbf{Top row: learn object slots for each frame.} Reading Figure~\ref{fig:overview}(a) from left to right, frozen DINO encodes image patches and the slot encoder groups their features. The decoder reconstructs current-frame features ($\mathcal L_{\rm feat}$) and predicts which patches in the next frame resemble each source patch ($\mathcal L_{\rm sim}$). The first target captures appearance and the second supplies a temporal correspondence signal. We then freeze the visual front end.

\textbf{Middle row: retain the cup push and account for a change of view.} Starting from the frozen slots $S_t,S_{t+\Delta}$ on the left, the inverse dynamics model (IDM) pools features with one query per hand and encodes their change as $z_t$. In the left half of the dashed box, keypoints guide where each query pools features ($\mathcal L_{\rm hand}$). A readout must recover both wrists' translation and relative rotation $a_t$ in the earlier head frame from $z_t$ alone ($\mathcal L_{\rm action}$). Omitting wrist motion incurs a readout error even if the predicted cup position is correct.

Following the same row to the right, a head turn can shift a stationary cup in the image. The forward dynamics model (FDM) receives $z_t$, current slots $S_t$ and measured head motion $n_t$, giving it a separate input to explain the shift. The predicted slots and their decoded features are compared with next-frame targets ($\mathcal L_{\rm fdm}$). The right half of the dashed box constrains the code itself: camera adversarial training discourages recoverable head motion and view consistency encourages similar codes before and after a synthetic warp.

\textbf{Bottom row: extract an action from two RGB frames.} Figure~\ref{fig:overview}(b) retains only the frozen DINO and slot encoder followed by the IDM with hand queries. The output is $z_t$; extraction requires neither pose annotations nor a forward prediction.

\subsection{Stage 1: Slot pretraining}
Slot pretraining uses frozen visual features as targets. DINOv2 divides each frame into patches. Slot attention~\cite{slotattention} pools patch features into $K$ slots. A predictor initializes the next frame’s slots from the current ones, then the new features refine them. The decoder has two targets for each patch: its current feature and a distribution over patches in the next frame.

A slot can reconstruct a cup in one frame without linking its features across frames. A second loss supplies that temporal signal. $\mathcal L_{\rm feat}$ is the mean squared error (MSE) against frozen DINO features. For a source patch $i$ on the cup rim, destination patches $j$ with similar frozen DINO keys receive more target probability $p_{ij}$. Several patches can share this mass. We train the predicted distribution $\widehat p_{ij}$ with cross-entropy~\cite{videosaur}:
\begin{equation}
\mathcal L_{\rm sim}=-\frac{1}{N}\sum_i\sum_j p_{ij}\log\widehat p_{ij}.
\end{equation}
Here $N$ is the number of source patches. The two losses train the slot grouping, predictor and decoder without hand labels. We retain this temporal objective throughout the slot experiments; its separate contribution was not measured.

\subsection{Stage 2: From slots to latent actions}
We freeze the pretrained visual front end and decoder. Each frame pair gives two sets of slots $S_t,S_{t+\Delta}\in\mathbb R^{K\times d_s}$ of width $d_s$.
The latent action encodes how features associated with the hands change between frames. The IDM processes both sets of slots jointly. Left- and right-hand queries pool the output features in each frame. Given the pooled hand features $q_t,q_{t+\Delta}$, a projection head flattens and maps their difference to
\begin{equation}
z_t=f_\theta(q_{t+\Delta}-q_t).
\end{equation}
Each query can pool context from the hand and the object it touches. We use $\operatorname{IDM}_\theta(S_t,S_{t+\Delta})$ to denote the full branch, including the queries and projection head.

\subsection{Accounting for head motion}
We measure both hand poses in the same frame of reference before computing a motion label. The relative camera pose maps the later hand pose into the earlier head coordinate frame. The translation and relative rotation of both hands form $a_t$. The camera's own translation and rotation form $n_t$. We normalize both targets using training-set statistics.

The FDM receives camera motion directly. It predicts future slots from the current slots, latent action and camera condition:
\begin{equation}
\widehat S_{t+\Delta}=\operatorname{FDM}(S_t,z_t,n_t).
\end{equation}
A learned token replaces the camera condition when the annotation is missing.

We compare the predicted slots with future slots and their decoded features with future DINO features $F_{t+\Delta}$. The prediction loss is
\begin{equation}
\begin{aligned}
\mathcal L_{\rm fdm}={}&\operatorname{MSE}(\operatorname{Decoder}(\widehat S_{t+\Delta}),F_{t+\Delta})\\
&+\lambda_{\rm slot}\operatorname{MSE}(\widehat S_{t+\Delta},S_{t+\Delta}),
\end{aligned}
\end{equation}
where $\lambda_{\rm slot}$ weights slot reconstruction. Gradients pass through the frozen decoder to train the FDM and action branch.

To reduce the camera motion recoverable from the code, the camera prediction head and the action encoder optimize the same prediction error in opposite directions. The head $g_{\rm cam}$ reads the code through a gradient reversal layer (GRL):
\begin{equation}
\widehat n_t=g_{\rm cam}(\operatorname{GRL}(z_t)),\quad
\mathcal L_{\rm adv}=\operatorname{MSE}_{\rm valid}(\widehat n_t,n_t).
\end{equation}
The camera head minimizes its error. The GRL acts as the identity on the forward pass and multiplies the gradient to the encoder by $-\gamma$ ($\gamma>0$)~\cite{grl}. The total loss thus includes $+\lambda_{\rm adv}\mathcal L_{\rm adv}$: the head gets the usual gradient, but the encoder gets $-\gamma\lambda_{\rm adv}\,\partial\mathcal L_{\rm adv}/\partial z_t$, discouraging predictable camera motion in the code.

View consistency encourages similar codes for the same action under an image warp. We apply a 2D transform to the later frame, encode the pair again to obtain $z'_t$, and use the original code as the target:
\begin{equation}
\mathcal L_{\rm inv}=\operatorname{MSE}(z'_t,\operatorname{sg}(z_t)).
\end{equation}
The stop-gradient operator $\operatorname{sg}$ blocks gradients to the original code. The transformed branch learns to match that code.

\subsection{Hand supervision and training}
Keypoints and poses from both frames supply two hand losses. One supervises the regions pooled by the queries. The other supervises the motion that can be read from the compressed code.

Keypoints tell each hand query where to pool information. For frame $\tau$, let $A_\tau$ be the query weights over slots, $M_\tau$ the slot masks, and $Y_\tau$ the binary target from hand keypoint heatmaps. On the same grid, the query response is $\widehat Y_\tau=A_\tau M_\tau$. Its loss is
\begin{equation}
\mathcal L_{\rm hand}=\tfrac12\sum_{\tau\in\{t,t+\Delta\}}
\operatorname{WBCE}_{\rm valid}(\widehat Y_\tau,Y_\tau),
\end{equation}
where WBCE is binary cross-entropy with a higher weight for positives. The response is a soft combination of slot masks, rather than an instance segmentation.

We supervise hand translation and rotation directly at the action bottleneck. A head $g_{\rm hand}$ predicts both hands' motion in the common frame from $z_t$:
\begin{equation}
\widehat a_t=g_{\rm hand}(z_t),\quad
\mathcal L_{\rm action}=\operatorname{MSE}_{\rm valid}(\widehat a_t,a_t).
\end{equation}
The region loss specifies where to pool features; the motion loss penalizes errors in what can be decoded from them. Section~\ref{sec:ablation-summary} tests the latter with a matched readout.

Joint training applies the hand-localization, motion-readout and camera constraints to the same future-prediction task. We add their weighted losses:
\begin{equation}
\begin{aligned}
\mathcal L={}&\lambda_{\rm fdm}\mathcal L_{\rm fdm}+\lambda_{\rm hand}\mathcal L_{\rm hand}
+\lambda_{\rm action}\mathcal L_{\rm action}\\
&+\lambda_{\rm adv}\mathcal L_{\rm adv}+\lambda_{\rm inv}\mathcal L_{\rm inv}.
\end{aligned}
\end{equation}
Each $\lambda$ sets the weight of one loss. The subscript $\mathrm{valid}$ restricts the average to the labels available for that loss. For example, Eva's end-effector poses provide motion supervision. Its missing hand heatmaps contribute to neither the numerator nor the denominator of the hand-region loss.

In this stage we train the IDM, hand queries, latent projection, FDM and readout heads. The visual front end and decoder remain frozen. The adversarial branch uses the GRL update above. We load cached slots for unperturbed frames and re-encode transformed images for the view-consistency branch.

\subsection{From video labels to robot commands}
\begin{figure}[!tb]\centering\includegraphics[width=.88\columnwidth]{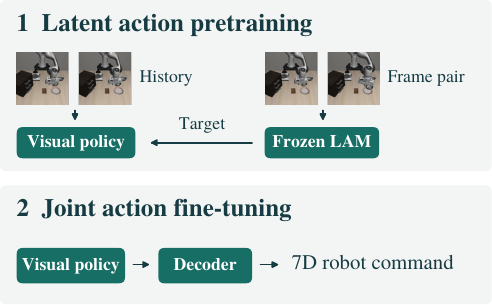}
\caption{\textbf{Policy training on a LIBERO-Spatial demonstration.} The frozen teacher reads frames 24 steps apart. The policy sees a history of three consecutive frames. In stage two, seven-dimensional robot commands supervise both the policy and the decoder. At deployment only observation history is needed.}
\label{fig:downstream}
\end{figure}

Figure~\ref{fig:downstream} shows how a two-frame label can train a policy that must act before the second frame arrives.
\par

We use the frozen extractor offline to label each training pair with $z_t$, following the latent-action policy training procedure of LAOM~\cite{laom}. Policy $\pi_\phi$ predicts this label from $o_{\leq t}$. For the bowl task we standardize the teacher labels per coordinate using training-set means and standard deviations, giving $\widetilde z_t$:
\begin{equation}
\widehat z_t=\pi_\phi(o_{\leq t}),\quad
\mathcal L_{\rm latent}=\ell_z(\widehat z_t,\operatorname{sg}(\widetilde z_t)).
\end{equation}

Expert demonstrations teach the policy how to turn its predicted code into a robot command. We jointly update the policy parameters $\phi$ and action decoder parameters $\psi$ against seven-dimensional commands:
\begin{equation}
\widehat u_t=D_\psi(\pi_\phi(o_{\leq t})),\quad
\mathcal L_{\rm ctrl}=\ell_a(\widehat u_t,u_t).
\end{equation}
Here $u_t$ is the expert robot command and $\widehat u_t$ the policy's prediction. Joint fine-tuning uses action MSE. We also test freezing the policy and fitting only the decoder. At deployment we evaluate $D_\psi(\pi_\phi(o_{\leq t}))$ without access to future frames.

\section{Egocentric diagnostics for LARYBench}\label{sec:egodiag}
A code can score well on LARYBench action recognition while retaining little decodable wrist motion~\cite{lary}. For example, a cup rising from a table may identify a lift even if the code omits which hand lifted it or how the wrist turned. In our results, LAOM reaches 70.56\% human action accuracy but its wrist-motion readout achieves only $R^2=0.0564$ on the same checkpoint (Tables~\ref{tab:1} and~\ref{tab:2}). Category recognition alone therefore does not tell us whether a code can label hand motion for imitation. Identifying which cue the model used requires an intervention. We use AD to measure recoverable motion and IC to measure responses to view or lighting changes at fixed action.
\setlength{\belowdisplayskip}{4pt plus 2pt minus 1pt}
\setlength{\belowdisplayshortskip}{4pt plus 1pt minus 1pt}

LARYBench asks whether a frozen code identifies an action category or predicts a robot command~\cite{lary}. Neither task separates two properties needed for egocentric imitation: retaining human wrist motion and resisting a change in appearance when the hand stays still. We test the first by decoding hand motion and the second by perturbing an image.

\subsection{Action decoding}
Two clips labeled as cup lifts can differ in which hand moves, how far it travels and how it turns. For action decoding (AD), an MLP predicts both hands’ translation and relative rotation in the common frame. We score $\widehat a_t^{\rm AD}$ against the measured $a_t$ with $R^2$. All models use the same unseen split and readout protocol. We average three readout seeds and retain negative scores. The unseen split holds out readout samples; some source videos also appear in the pretraining pool.

\subsection{Viewpoint and lighting interventions}
Actions and viewing conditions can change together in natural videos. Classification and regression scores alone cannot isolate viewpoint or lighting. We hold the action fixed and transform only the second image. For sample $i$, identical frames give $z_i^{\rm id}=E(o_t,o_t)$, a real transition gives $z_i^{\rm real}=E(o_t,o_{t+\Delta})$, and an intervention gives $z_i^\kappa=E(o_t,T_\kappa(o_t))$. Here $\kappa\in\{\mathrm{camera},\mathrm{photo}\}$ denotes a 2D warp or photometric change. The identical-frame baseline allows for nonzero codes even without motion.

\begin{figure}[!tb]\centering\includegraphics[width=.89\columnwidth]{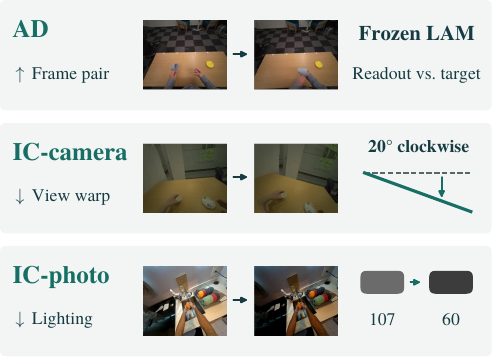}
\caption{\textbf{A frozen code is used for both action decoding and interventions.} AD compares the readout prediction with measured hand motion. The figure rotates the frame clockwise by $20^\circ$ for illustration. The evaluated 2D warp for this sample included a $4.6^\circ$ rotation. The lighting change reduces mean grayscale intensity from 107 to 60 (range 0--255). Each IC divides the median intervention response by the median real-transition response, both measured from identical input frames.}
\label{fig:diagnostics}
\end{figure}

Intervention sensitivity (IC) measures the change in the code relative to its change on a real transition. For $u\in\{\mathrm{camera},\mathrm{photo},\mathrm{real}\}$, let $d_i^u=\lVert z_i^u-z_i^{\rm id}\rVert_s$. To compute this norm, we divide coordinate $j$ by $s_j+10^{-6}$ and take the Euclidean norm. Here $s_j$ is the standard deviation of that coordinate across real transitions for the same model on the fixed cohort. We compute
\begin{equation}
\mathrm{IC}_\kappa=\frac{\operatorname{median}_i d_i^\kappa}{\operatorname{median}_i d_i^{\rm real}},
\quad \kappa\in\{\mathrm{camera},\mathrm{photo}\}.
\end{equation}
Lower IC means a smaller response to the intervention relative to real motion. The ratio is undefined for a zero denominator. Figure~\ref{fig:diagnostics} shows the inputs used for each test.

IC must be interpreted alongside AD: a code that responds weakly to visual changes is useful for imitation only if it still retains hand motion. IC-camera measures sensitivity to a synthetic warp. Section~5.2 examines real turns matched for one wrist; isolating rotation also requires comparable head translation.

\section{Experiments}
We test what frozen codes retain, how their labels guide robot policies, and which components preserve wrist motion or reduce sensitivity to viewpoint.

\begin{figure}[!tb]\centering\includegraphics[width=.96\columnwidth]{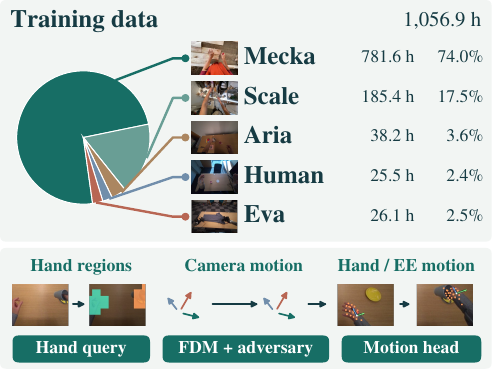}
\caption{\textbf{Mecka accounts for 74\% of the latent action training pool.} Hand keypoints supervise the queries, hand or end-effector motion supervises the action code, and camera motion conditions forward prediction and supervises the adversary. The LAM pool spans about 1,056.9 hours on the sampling grid, which can count source video more than once. The slot encoder uses a separate 1,120.2-hour pool. Eva supplies end-effector poses but no hand heatmaps. We assume a fixed camera for Eva.}
\label{fig:data}
\end{figure}

\paragraph{Data and comparison settings.}
All eight models in Table~\ref{tab:1} trained for 56,000 updates on the same pool of mostly human egocentric footage and some robot demonstrations (Figure~\ref{fig:data}). We used existing egocentric recordings and their accompanying hand and camera annotations from EgoVerse~\cite{egoverse} to train and evaluate EgoLAM. No participants were recruited and no new human demonstrations were recorded for this study. Each retains its native code shape and supervision: action decoding for LAOM and CLAM~\cite{laom,clam}, visual tokens for LAPA and Moto~\cite{lapa,moto}, a geometry teacher for GeoLAM~\cite{geolam}, and frozen DINO feature targets for the two LARYBench LAPA variants~\cite{lary,dinov2,dinov3}. In Table~S3, 128 dimensions yield the best classification and CALVIN scores, and 256 the best RoboCOIN and AgiBot scores. Both yield 0.132 VLABench MSE at the reported precision. Table~\ref{tab:core-ablation} tests individual components.

\begin{figure}[!tb]\centering\includegraphics[width=.88\columnwidth]{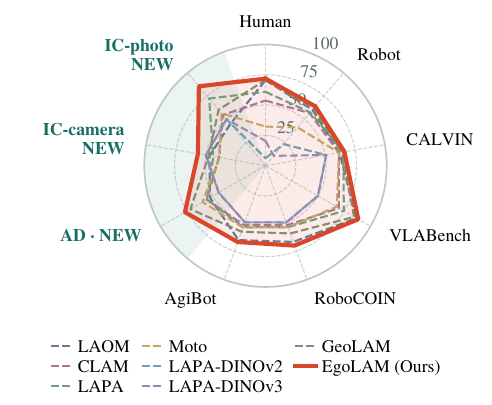}
\caption{\textbf{Original tasks and new diagnostics for fixed checkpoints.} The nine axes show the individual scores. The three dark green axes marked NEW give unseen AD (MLP), IC-camera and IC-photo. The red curve uses the same EgoLAM checkpoint throughout. Larger radii are better: top-1 accuracy (\%), $100/(1+\mathrm{MSE})$, $50(1+R^2)$ for AD, and $100/(1+\mathrm{IC})$ for IC. Tables~\ref{tab:1} and~\ref{tab:2} give the raw values.}
\label{fig:radar}
\end{figure}

\begin{table*}[!t]\centering
\captionof{table}{\textbf{Action classification and robot command regression on LARYBench.} We report top-1 accuracy (\%) for human and robot actions, and normalized MSE on CALVIN-5, VLABench-5, RoboCOIN-10 (seen) and AgiBot-45 (seen). Bold marks the best value in each column, including ties at the displayed precision.}
\label{tab:1}
\centering\begingroup\fontsize{8}{10}\selectfont\setlength{\tabcolsep}{2pt}\renewcommand{\arraystretch}{1.1}
\begin{tabularx}{\linewidth}{@{}>{\raggedright\arraybackslash}p{34mm}*{9}{Y}@{}}
\toprule
\textbf{Model} & \textbf{\shortstack{Latent\\shape}} & \textbf{\shortstack{Human\\Top-1 $\uparrow$}} & \textbf{\shortstack{Robot\\Top-1 $\uparrow$}} & \textbf{\shortstack{CALVIN\\MSE $\downarrow$}} & \textbf{\shortstack{VLA\\MSE $\downarrow$}} & \textbf{\shortstack{Robo\\MSE $\downarrow$}} & \textbf{\shortstack{AgiBot\\MSE $\downarrow$}} & \textbf{\shortstack{Mean acc.\\$\uparrow$}} & \textbf{\shortstack{Mean MSE\\$\downarrow$}} \\
\midrule
LAOM & 1$\times$256 & 70.56 & 60.19 & 0.616 & 0.156 & 0.495 & 0.532 & 65.37 & 0.450 \\
CLAM & 1$\times$8 & 53.76 & 56.65 & 0.626 & 0.430 & 0.916 & 0.917 & 55.20 & 0.722 \\
LAPA & 4$\times$1024 & 60.95 & 57.39 & 0.533 & 0.341 & 0.851 & 0.858 & 59.17 & 0.646 \\
Moto & 8$\times$32 & 32.28 & 41.25 & 0.623 & 0.475 & 0.856 & 0.842 & 36.76 & 0.699 \\
LAPA-DINOv2 & 16$\times$1024 & 6.23 & 23.25 & 0.970 & 1.006 & 1.006 & 1.013 & 14.74 & 0.999 \\
LAPA-DINOv3 & 16$\times$1024 & 20.43 & 10.61 & 0.964 & 1.004 & 1.006 & 1.014 & 15.52 & 0.997 \\
GeoLAM & 4$\times$32 & 70.71 & 61.41 & 0.563 & 0.185 & 0.681 & 0.727 & 66.06 & 0.539 \\
\midrule
EgoLAM & 1$\times$128 & \textbf{71.81} & \textbf{63.88} & \textbf{0.513} & \textbf{0.132} & \textbf{0.426} & \textbf{0.493} & \textbf{67.85} & \textbf{0.391} \\
\bottomrule
\end{tabularx}

\endgroup\par\medskip
\end{table*}
\begin{figure*}[!t]\centering\input{tables/metrics-panel}\par\medskip
\begin{minipage}[t]{.445\linewidth}
\captionof{table}{\textbf{Hand-motion decoding and visual sensitivity.}}
\label{tab:2}
\centering\begingroup\fontsize{9}{10}\selectfont\setlength{\tabcolsep}{2pt}\renewcommand{\arraystretch}{1.23}
\begin{tabularx}{\linewidth}{@{}lc*{2}{Y}@{}}
\toprule
 & Unseen AD & \multicolumn{2}{c}{Visual sensitivity (IC)} \\
\cmidrule(lr){2-2}\cmidrule(l){3-4}
\textbf{Model} & $R^2\uparrow$ & Camera $\downarrow$ & Photo $\downarrow$ \\
\midrule
LAOM & 0.0564 $\pm$ 0.0027 & 1.1323 & 1.2216 \\
CLAM & 0.1170 $\pm$ 0.0012 & 1.5578 & 0.8292 \\
LAPA & 0.1140 $\pm$ 0.0013 & 1.2885 & 0.3821 \\
Moto & 0.1983 $\pm$ 0.0004 & 1.5978 & 0.7727 \\
LAPA-DINOv2 & -0.1118 $\pm$ 0.0018 & 1.0018 & 1.0014 \\
LAPA-DINOv3 & -0.1068 $\pm$ 0.0021 & 1.0032 & 0.9987 \\
GeoLAM & 0.4312 $\pm$ 0.0030 & 1.1048 & 0.6606 \\\midrule
\textbf{EgoLAM} & \textbf{0.5343 $\pm$ 0.0018} & \textbf{0.7673} & \textbf{0.1699} \\
\bottomrule
\end{tabularx}

\endgroup\tablenote{AD reports mean $\pm$ SD across three readout seeds on the unseen split. Both IC scores use the same fixed pairs. Bold marks the best value in each column.}\end{minipage}\hfill\begin{minipage}[t]{.535\linewidth}\centering
\captionof{table}{\textbf{Closed-loop success on LIBERO.}}\label{tab:3}
\begingroup\fontsize{8.5}{10}\selectfont\setlength{\tabcolsep}{0pt}\renewcommand{\arraystretch}{1.04}
\begin{tabular*}{\linewidth}{@{\extracolsep{\fill}}lrr@{\hskip 5pt}rr@{\hskip 5pt}rr@{}}
\toprule
 & \multicolumn{2}{c}{Spatial} & \multicolumn{2}{c}{Object} & \multicolumn{2}{c}{Goal} \\
\cmidrule(lr){2-3}\cmidrule(lr){4-5}\cmidrule(l){6-7}
\textbf{Policy} & SR & $\Delta$ & SR & $\Delta$ & SR & $\Delta$ \\
\midrule
Expert BC & $59.27$ & $0.00$ & $41.07$ & $0.00$ & $51.60$ & $0.00$ \\
\midrule
\textbf{EgoLAM} & $\mathbf{81.40}$ & $\mathbf{+22.13}$ & $\mathbf{77.33}$ & $\mathbf{+36.27}$ & $\mathbf{79.40}$ & $\mathbf{+27.80}$ \\
LAOM & $61.33$ & $+2.07$ & $43.87$ & $+2.80$ & $48.73$ & $-2.87$ \\
CLAM & $63.73$ & $+4.47$ & $38.53$ & $-2.53$ & $62.27$ & $+10.67$ \\
LAPA & $56.33$ & $-2.93$ & $46.47$ & $+5.40$ & $54.13$ & $+2.53$ \\
Moto & $51.33$ & $-7.93$ & $45.87$ & $+4.80$ & $59.93$ & $+8.33$ \\
LAPA-DINOv2 & $58.47$ & $-0.80$ & $53.40$ & $+12.33$ & $67.80$ & $+16.20$ \\
LAPA-DINOv3 & $57.40$ & $-1.87$ & $46.87$ & $+5.80$ & $\underline{68.67}$ & $+17.07$ \\
GeoLAM & $\underline{74.33}$ & $+15.07$ & $\underline{65.87}$ & $+24.80$ & $67.47$ & $+15.87$ \\
\midrule
EgoLAM gain & \multicolumn{2}{c}{$+7.07$} & \multicolumn{2}{c}{$+11.47$} & \multicolumn{2}{c}{$+10.73$} \\
\bottomrule
\end{tabular*}\endgroup
\tablenote{SR is success (\%). $\Delta$ is the gain over expert BC (pp). Bold marks the best result. Underlines mark the strongest other LAM, the reference for the last row (pp).}
\end{minipage}\par\medskip\centering\par\smallskip
\end{figure*}

\begin{figure*}[!t]\centering\begin{minipage}[t]{.90\textwidth}\vspace{0pt}\centering
\includegraphics[width=\linewidth]{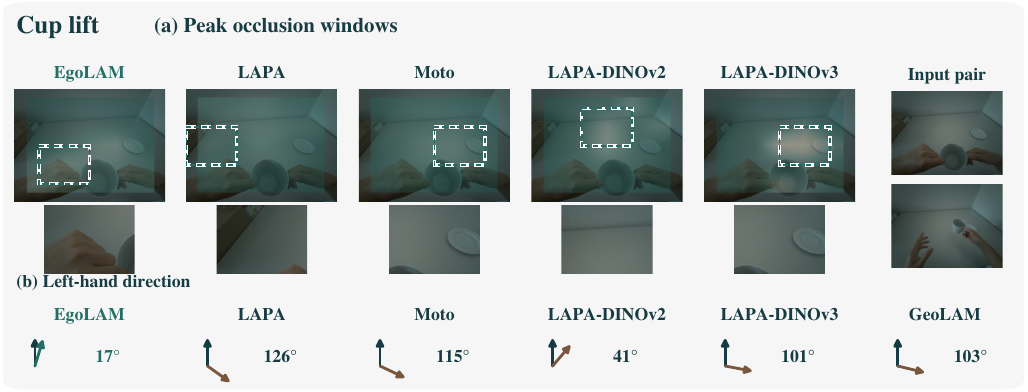}
\captionof{figure}{\textbf{On this cup lift, EgoLAM responds most to occluding the acting hand and recovers its displacement direction with the smallest error ($17^\circ$).} (a) A window spanning one third of each image dimension masks both frames at $5\times5$ locations. Boxes and crops mark peak responses; colors give each window’s share of the total for that model. (b) With the measured direction drawn upright, the colored arrow shows the 3D angular error (same pair, seed 42).}
\label{fig:cup-evidence}
\end{minipage}\end{figure*}
\begin{figure}[!tb]\centering\includegraphics[width=.95\columnwidth]{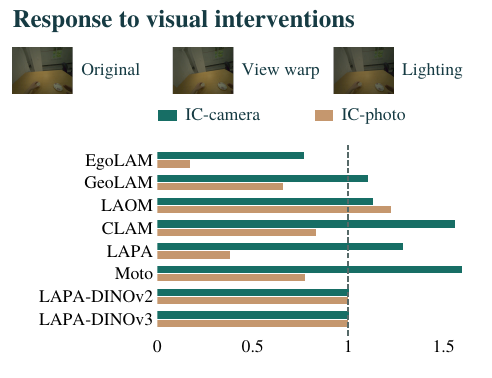}
\caption{\textbf{EgoLAM has the lowest relative response to both interventions among eight models. View warps remain more disruptive than lighting changes.} Across 605 pairs, median responses to a synthetic warp and to lighting are 0.767 and 0.170 times the median real-transition response, respectively. The dashed line marks equality. Together with AD (Table~\ref{tab:2}), these results show that low intervention sensitivity coexists with decodable wrist motion.}
\label{fig:ic-cohort}
\end{figure}

\subsection{LARYBench}

\label{sec:lary}
Figure~\ref{fig:radar} shows how the two original LARYBench task families relate to the three egocentric diagnostics.
LARYBench tests whether one frozen representation supports both action classification and continuous robot command regression. We classify human and robot actions and evaluate regression on CALVIN-5, VLABench-5, RoboCOIN-10 (seen) and AgiBot-45 (seen). Readouts train on one GPU at global batch size 256 for 20 classification or 50 regression epochs. We report top-1 accuracy and normalized MSE, with separate macro averages for complete task groups.

Among the eight models trained here, EgoLAM has the highest mean accuracy (67.85\%) and lowest mean MSE (0.391). LAOM scores 65.37\% and 0.450, and GeoLAM 66.06\% and 0.539 (Table~\ref{tab:1}).

\subsection{LARYBench extensions}
AD readouts train for 20 epochs (global batch 2,048). Table~\ref{tab:2} gives held-out means and sample standard deviations over seeds 42--44 for each frozen checkpoint. Both IC tests use the same 605 pairs.

EgoLAM preserves more decodable hand motion and responds less to controlled visual changes than GeoLAM and LAOM. Their AD scores are 0.5343, 0.4312 and 0.0564, respectively. EgoLAM scores 0.7673 on IC-camera and 0.1699 on IC-photo (Table~\ref{tab:2}).

On clips with large real head turns, EgoLAM decodes wrist displacement more accurately than LAPA. Matching one wrist’s motion within each held-out video gives similar errors for small turns (2.73 vs.\ 2.83 cm) but a larger gap for large turns (5.47 vs.\ 9.63 cm). Validation agrees, supporting the motivation for direct wrist supervision. Head translation also differs between groups. The supplement details the matching procedure, results and interpretation for this comparison (Tables~S5--S6).

In the cup lift (Figure~\ref{fig:cup-evidence}), EgoLAM responds most to occluding the acting left hand. Its readout recovers that hand’s displacement direction to within $17^\circ$, compared with $41^\circ$--$126^\circ$ for the five other readouts.

\subsection{Downstream tasks}\label{sec:downstream-eval}
We compare latent action teachers within a common policy-training protocol on LIBERO~\cite{libero}. For each task, all teachers label the same 24-frame-gap image pairs from the same 40 LIBERO training demonstrations. The policy first predicts these frozen teacher codes for 10 epochs, then jointly fine-tunes with its action decoder on expert commands from the same trajectories for 2,500 updates. Direct BC uses the same action samples and updates. Table~\ref{tab:3} reports success and gain over BC in percentage points across ten tasks per suite, three seeds per task and 50 initial states per seed.

EgoLAM leads on Spatial, Object and Goal with 81.40\%, 77.33\% and 79.40\% success. Direct BC scores 59.27\%, 41.07\% and 51.60\%, giving gains of 22.13, 36.27 and 27.80 points. The margins over GeoLAM on Spatial and Object and LAPA-DINOv3 on Goal are 7.07, 11.47 and 10.73 points, respectively (Table~\ref{tab:3}).

Measured wrist motion supervises EgoLAM. Its predicted wrist translations and rotations can then serve as targets for policy pretraining. On two Spatial bowl tasks with matched demonstrations, targeting $g_{\rm hand}(z_t)$ yielded 76.67\% mean success after fine-tuning, 9.00 points above the 67.67\% obtained with the same teacher’s full $z_t$ (Table~S7).

\subsection{Full extractor size and throughput with cached features}\label{sec:efficiency}
Figure~\ref{fig:efficiency} times single-pair extraction on one Xeon Platinum 8457C thread (FP32). Cached features serve overlapping pairs, as in EgoLAM training; CLAM starts from preprocessed images. We verify outputs and time at least three calls after warmup. Counts include the full RGB-to-action path.

EgoLAM has 90.89M parameters (89.54M front end, 1.35M action branch), third smallest after CLAM (2.69M) and LAOM (12.07M). Its advantage is median single-pair latency: 0.85 ms with cached slots versus LAOM's 1.47 ms with cached endpoint-stack features. At batch size 32, LAOM is faster (8.42 vs.\ 10.50 ms).

\medskip
\subsection{Component ablations}\label{sec:ablation-summary}
Table~\ref{tab:core-ablation} shows that removing wrist supervision or replacing slots with a fixed grid lowers AD. Removing view consistency raises IC-camera above one: the median response to a synthetic warp exceeds that to a real transition. Removing the camera adversary or FDM camera input also raises IC-camera. Supplementary Tables~S1--S4 report all ablations.

\par\medskip\noindent\begin{minipage}{\columnwidth}\centering
\captionof{table}{Single-factor ablations. The wrist-loss control is its paired original-head model (AD 0.5592). The other rows use the main model as their control (AD 0.5343 for the grid comparison).}\label{tab:core-ablation}
\small\setlength{\tabcolsep}{2pt}
\begin{tabular*}{\linewidth}{@{}l@{\extracolsep{\fill}}lrr@{}}\toprule
Change & Metric & Control & Ablated \\ \midrule
Remove hand-motion loss & AD $\uparrow$ & \textbf{0.5592} & 0.3502 \\
Replace slots with a grid & AD $\uparrow$ & \textbf{0.5343} & 0.4823 \\
Remove view consistency & IC-camera $\downarrow$ & \textbf{0.7673} & 1.1276 \\
Remove camera adversary & IC-camera $\downarrow$ & \textbf{0.7673} & 0.8420 \\
Omit FDM camera input & IC-camera $\downarrow$ & \textbf{0.7673} & 0.8245 \\
\bottomrule\end{tabular*}\end{minipage}\par

\section{Discussion and conclusion}
A latent action for imitation must retain wrist motion even when the camera moves. EgoLAM uses hand poses to define this motion in a common frame and gives camera motion to the forward model separately. The paired motion-loss ablation supports the first choice, and the view-consistency ablation shows that training with warps reduces sensitivity to that intervention. The full model had the highest hand-motion decoding score among the eight evaluated checkpoints.

After expert fine-tuning, mean success on three LIBERO suites reached 79.38\%, a gain of 28.73 points over direct BC and 9.76 points over the strongest competing latent action teacher per suite, on average.

EgoLAM could supply pretraining labels for first-person demonstrations without robot commands. Consider a wearer who looks toward a shelf while lifting a cup. On independent head-mounted videos, we would match wrist motion to isolate the effect of real head turns, then fine-tune policies on matched demonstrations and measure real pick-and-place success.

\clearpage
\section*{LLM Usage Statement}
We used OpenAI Codex to help draft and polish the manuscript and summarize the results. The authors take full responsibility for all reported results.\par\medskip
{\small\bibliographystyle{ieeenat_fullname}\bibliography{references}}

\onecolumn
\phantomsection\pdfbookmark[0]{Supplementary Material}{egolam-supplement}
\begin{center}{\Large\bfseries Supplementary Material}\end{center}
\medskip
\appendix
\setcounter{table}{0}\renewcommand{\thetable}{S\arabic{table}}
\renewcommand{\theHtable}{supp.\arabic{table}}
\captionsetup[table]{skip=3pt}
\noindent\begin{minipage}[t]{.49\textwidth}\vspace{0pt}\centering
\captionof{table}{Direct hand-motion supervision}\label{tab:motion}
\begingroup\small\setlength{\tabcolsep}{2pt}\renewcommand{\arraystretch}{1.04}\begin{tabularx}{\linewidth}{@{}>{\raggedright\arraybackslash}p{.40\linewidth}*{3}{Y}@{}}\toprule
Setting & Mean AD $R^2\uparrow$ & IC-camera $\downarrow$ & IC-photo $\downarrow$ \\ \midrule
Original head (paired control) & \textbf{0.5592} & \textbf{0.7032} & \textbf{0.1578} \\
Without direct motion supervision & 0.3502 & 0.8235 & 0.1965 \\
\bottomrule\end{tabularx}\endgroup\end{minipage}\hfill\begin{minipage}[t]{.49\textwidth}\vspace{0pt}\centering
\captionof{table}{Latent action head}\label{tab:head}
\begingroup\small\setlength{\tabcolsep}{2pt}\renewcommand{\arraystretch}{1.04}\begin{tabularx}{\linewidth}{@{}>{\raggedright\arraybackslash}p{.40\linewidth}*{3}{Y}@{}}\toprule
Setting & Mean AD $R^2\uparrow$ & IC-camera $\downarrow$ & IC-photo $\downarrow$ \\ \midrule
Original head (paired control) & \textbf{0.5592} & \textbf{0.7032} & \textbf{0.1578} \\
Residual head & 0.5020 & 0.8757 & 0.1706 \\
\bottomrule\end{tabularx}\endgroup\end{minipage}\par\smallskip\noindent\begin{minipage}{\textwidth}\centering
\captionof{table}{Latent action dimension}\label{tab:dim}
\begingroup\small\setlength{\tabcolsep}{3pt}\renewcommand{\arraystretch}{1.04}\begin{tabularx}{\linewidth}{@{}l*{6}{Y}@{}}\toprule
Dimension & Human $\uparrow$ & Robot $\uparrow$ & CALVIN $\downarrow$ & VLA $\downarrow$ & Robo $\downarrow$ & AgiBot $\downarrow$ \\ \midrule
16 & 60.25 & 48.98 & 0.806 & 0.542 & 0.891 & 0.929 \\
32 & 69.76 & 61.49 & 0.619 & 0.250 & 0.628 & 0.700 \\
64 & 71.28 & 63.40 & 0.639 & 0.165 & 0.469 & 0.543 \\
128 (baseline) & \textbf{71.81} & \textbf{63.88} & \textbf{0.513} & \textbf{0.132} & 0.426 & 0.493 \\
256 & 70.57 & 63.51 & 0.572 & \textbf{0.132} & \textbf{0.417} & \textbf{0.477} \\
512 & 69.65 & 62.99 & 0.5827 & 0.158 & 0.539 & 0.621 \\
\bottomrule\end{tabularx}\endgroup\end{minipage}\par\smallskip\noindent\begin{minipage}{\textwidth}\centering
\captionof{table}{Supervision and model components}\label{tab:loss-components}
\begingroup\small\setlength{\tabcolsep}{4pt}\renewcommand{\arraystretch}{1.06}
\begin{tabularx}{\linewidth}{@{}l>{\raggedright\arraybackslash}Xccc@{}}\toprule
Group & Change from baseline & Mean AD $R^2\uparrow$ & IC-camera $\downarrow$ & IC-photo $\downarrow$ \\ \midrule
Baseline & EgoLAM (128 dimensions, full configuration) & \textbf{0.5343} & \textbf{0.7673} & \textbf{0.1699} \\
\midrule
Loss & Remove hand-region loss, keep motion readout & 0.5079 & 0.8299 & 0.1769 \\
 & Remove adversarial loss, keep FDM camera input & 0.4660 & 0.8420 & 0.2300 \\
 & Remove view consistency on warped frames & 0.5323 & 1.1276 & 0.2682 \\
\midrule
Component & Replace FDM camera input with the missing-condition token, keep aligned hand labels and adversarial loss & 0.2242 & 0.8245 & 0.3598 \\
 & Replace slots with fixed spatial grid features, keeping the backbone and supervision & 0.4823 & 0.9077 & 0.2080 \\
\bottomrule\end{tabularx}\endgroup\end{minipage}\par

\section*{Supplementary experiment: wrist decoding with EgoLAM and LAPA during real head turns}
\textbf{Samples and readouts.} We focus on LAPA~\cite{lapa}, the pixel-reconstruction baseline discussed in the introduction. This exploratory comparison asks how wrist-decoding errors change for the two complete models during real head turns in EgoVerse manipulation videos. EgoLAM and LAPA each use their own frozen seed-42 wrist MLP across both groups and splits. Head turn is the net 3D rotation between measured head poses 15 frames apart. We require valid head poses and tracking for both wrists. Of 7,547 eligible held-out frame pairs, 5,614 have small turns ($\leq5^\circ$) and 433 have large turns ($\geq15^\circ$). The corresponding validation counts are 7,334, 5,589 and 320.

\textbf{Matching.} The original design required the same video and task, displacement vectors within 2 cm and relative rotations within $10^\circ$ for each wrist, and head translation vectors within 2 cm. Neither split yielded a match. We therefore conducted an exploratory analysis matching one wrist at a time with the same wrist tolerances, allowing the other hand and head translation to differ. We matched nonoverlapping windows one-to-one without replacement within each hand, first maximizing the number of pairs and then minimizing normalized distance, without using prediction errors. Both models share the resulting 40 wrist pairs (33 left, 7 right; 16 episodes) on the held-out split and 33 pairs (26 left, 7 right; 19 separate episodes) on validation.
\par\medskip\noindent\begin{minipage}{\textwidth}\centering\captionof{table}{Mean wrist error during small and large real head turns. Translation is Euclidean displacement error; rotation is the 3D geodesic error. Each matched wrist contributes equally. Bold marks the lowest error in each column.}\label{tab:turn-errors}
\small\setlength{\tabcolsep}{3pt}\renewcommand{\arraystretch}{1.10}
\begin{tabular*}{\linewidth}{@{}l@{\extracolsep{\fill}}rrrrrrrr@{}}\toprule
 & \multicolumn{4}{c}{Held-out} & \multicolumn{4}{c}{Validation}\\\cmidrule(lr){2-5}\cmidrule(lr){6-9}
 & \multicolumn{2}{c}{Translation (cm)} & \multicolumn{2}{c}{Rotation (deg)} & \multicolumn{2}{c}{Translation (cm)} & \multicolumn{2}{c}{Rotation (deg)}\\
Model & Small & Large & Small & Large & Small & Large & Small & Large\\\midrule
EgoLAM & \textbf{2.73} & \textbf{5.47} & \textbf{7.79} & 11.14 & \textbf{2.49} & \textbf{5.90} & 8.28 & \textbf{10.82} \\
LAPA & 2.83 & 9.63 & 8.25 & \textbf{10.35} & 2.50 & 9.25 & \textbf{7.15} & 10.85 \\
\bottomrule\end{tabular*}
\end{minipage}\par
\par\medskip\noindent\begin{minipage}{\textwidth}\centering\captionof{table}{Error increase (large minus small) on the same matched wrists. Bold marks the smaller increase in each row. The last column gives LAPA’s increase minus EgoLAM’s, with exploratory 95\% intervals from 10,000 joint bootstrap draws over episodes stratified by source, without multiplicity correction.}\label{tab:turn-contrasts}
\small\setlength{\tabcolsep}{4pt}\renewcommand{\arraystretch}{1.10}
\begin{tabular*}{\linewidth}{@{}ll@{\extracolsep{\fill}}rrc@{}}\toprule
Split & Error metric & EgoLAM & LAPA & Difference [95\% interval]\\\midrule
Held-out & Translation (cm) & \textbf{+2.75} & +6.80 & $+4.06\;[+2.08, +6.11]$ \\
 & Rotation (deg) & +3.35 & \textbf{+2.10} & $-1.25\;[-3.77, +1.15]$ \\
Validation & Translation (cm) & \textbf{+3.40} & +6.75 & $+3.35\;[+1.16, +5.56]$ \\
 & Rotation (deg) & \textbf{+2.54} & +3.71 & $+1.17\;[-2.04, +3.56]$ \\
\bottomrule\end{tabular*}
\end{minipage}\par

\textbf{Interpretation.} The increase in translation error is smaller for EgoLAM than LAPA by 4.06 cm on the held-out split and 3.35 cm on validation. The rotation contrasts have opposite signs across splits, and both intervals include zero. The evidence for an EgoLAM advantage over LAPA is therefore specific to wrist translation. Median head translation increases from 1.28 to 6.39 cm across turn groups on the held-out split and from 1.08 to 5.92 cm on validation. Requiring head translation vectors within 2 cm leaves only two wrist pairs (one episode) and zero pairs, respectively. These comparisons characterize large-turn scenes; they cannot isolate an effect of rotation alone.

\par\medskip\noindent\textbf{Which teacher target should the policy imitate?}
Frozen EgoLAM labels the same frame pairs (gap 24) with $z_t$ or its wrist prediction $g_{\rm hand}(z_t)$. Both targets use training-set standardization. A fixed map preserves mean squared distances while embedding the 12 wrist coordinates in the same 128-dimensional policy interface. Both policy variants use the same 40 training demonstrations per task, policy backbone and action decoder. After 10 epochs of target prediction, each policy and its decoder are jointly fine-tuned for 2,500 updates on expert commands from the same demonstrations. Evaluation uses the same 50 initial states per seed.
\par\smallskip\noindent\begin{minipage}{\textwidth}\centering
\captionof{table}{Closed-loop success on two bowl-to-plate tasks. Counts pool three seeds with 50 starts each. Bold marks the higher rate; differences are full minus wrist in percentage points.}\label{tab:policy-target}
\small\setlength{\tabcolsep}{4pt}\renewcommand{\arraystretch}{1.12}
\begin{tabular*}{\linewidth}{@{}l@{\extracolsep{\fill}}rrrrr@{}}\toprule
 & \multicolumn{2}{c}{Full $z_t$} & \multicolumn{2}{c}{Wrist target $g_{\rm hand}(z_t)$} & \\
\cmidrule(lr){2-3}\cmidrule(lr){4-5}
Bowl location & Successes/trials & Success rate & Successes/trials & Success rate & Difference (pp)\\\midrule
Table center & 126/150 & 84.00\% & 142/150 & \textbf{94.67\%} & $-10.67$\\
Next to plate & 77/150 & 51.33\% & 88/150 & \textbf{58.67\%} & $-7.33$\\\midrule
Equal-task mean & 203/300 & 67.67\% & 230/300 & \textbf{76.67\%} & $-9.00$\\\bottomrule
\end{tabular*}\end{minipage}\par

\end{document}